\documentclass[letterpaper, 10pt, conference]{ieeeconf}      

\IEEEoverridecommandlockouts                              

\usepackage{amsmath} 
\usepackage{amssymb}  
\usepackage[rgb,table]{xcolor}
\usepackage{graphicx}
\usepackage{gensymb}
\usepackage{booktabs}
\usepackage{multirow}
\usepackage{array}
\usepackage{siunitx}
\usepackage{import}
\usepackage{hyperref}
\usepackage[english]{babel}
\addto\extrasenglish{%
}

\title{\LARGE \bf
Adaptive Model-Based Control of Quadrupeds via Online System Identification using Kalman Filter
}

\author{Jonas Haack$^{1, 2}$, Franek Stark$^{1}$, Shubham Vyas$^{1, 2}$, Frank Kirchner$^{1, 2}$, Shivesh Kumar$^{1, 3}$
\thanks{\textcopyright 2025 IEEE.  Personal use of this material is permitted.  Permission from IEEE must be obtained for all other uses, in any current or future media, including reprinting/republishing this material for advertising or promotional purposes, creating new collective works, for resale or redistribution to servers or lists, or reuse of any copyrighted component of this work in other works. Part of this work was supported by AAPLE (grant number 50WK2275) funded by the German Federal Ministry for Economic Affairs and Climate Action (BMWK), and CoEx (grant number 01IW24008) funded by the German Federal Ministry for Education and Research, and is partially supported with funds from the federal state of Bremen for setting up the Underactuated Robotics Lab (265/004-08-02-02-30365/2024-102966/2024-740847/2024).}
\thanks{$^{1}$ Robotics Innovation Center, German Research Center for Artificial Intelligence (DFKI GmbH), Bremen, Germany
{\tt\small jonas.haack@dfki.de}}%
\thanks{$^{2}$ AG Robotik, University of Bremen, Bremen, Germany}%
\thanks{$^{3}$ Dynamics Division, Department of Mechanics \& Maritime Sciences, Chalmers University of Technology, Gothenburg, Sweden}%
}

\begin{document}

\maketitle
\thispagestyle{empty}
\pagestyle{empty}

\begin{abstract}

Many real-world applications require legged robots to be able to carry variable payloads. Model-based controllers such as model predictive control (MPC) have become the de facto standard in research for controlling these systems. However, most model-based control architectures use fixed plant models, which limits their applicability to different tasks. In this paper, we present a Kalman filter (KF) formulation for online identification of the mass and center of mass (COM) of a four-legged robot. We evaluate our method on a quadrupedal robot carrying various payloads and find that it is more robust to strong measurement noise than classical recursive least squares (RLS) methods. Moreover, it improves the tracking performance of the model-based controller with varying payloads when the model parameters are adjusted at runtime.
\end{abstract}
\section{Introduction}

\begin{figure}[t]
    \centering
    \def\svgwidth{\columnwidth}
\begingroup%
  \makeatletter%
  \providecommand\color[2][]{%
    \errmessage{(Inkscape) Color is used for the text in Inkscape, but the package 'color.sty' is not loaded}%
    \renewcommand\color[2][]{}%
  }%
  \providecommand\transparent[1]{%
    \errmessage{(Inkscape) Transparency is used (non-zero) for the text in Inkscape, but the package 'transparent.sty' is not loaded}%
    \renewcommand\transparent[1]{}%
  }%
  \providecommand\rotatebox[2]{#2}%
  \newcommand*\fsize{\dimexpr\f@size pt\relax}%
  \newcommand*\lineheight[1]{\fontsize{\fsize}{#1\fsize}\selectfont}%
  \ifx\svgwidth\undefined%
    \setlength{\unitlength}{252.00820541bp}%
    \ifx\svgscale\undefined%
      \relax%
    \else%
      \setlength{\unitlength}{\unitlength * \real{\svgscale}}%
    \fi%
  \else%
    \setlength{\unitlength}{\svgwidth}%
  \fi%
  \global\let\svgwidth\undefined%
  \global\let\svgscale\undefined%
  \makeatother%
  \begin{picture}(1,0.68923606)%
    \lineheight{1}%
    \setlength\tabcolsep{0pt}%
    \put(0,0){\includegraphics[width=\unitlength,page=1]{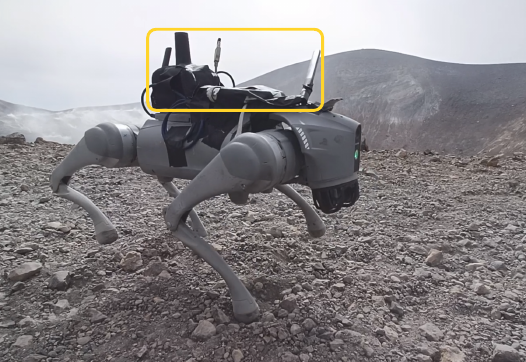}}%
    \put(0.11474892,0.57115852){\color[rgb]{0,0,0}\makebox(0,0)[lt]{\lineheight{1.25}\smash{\begin{tabular}[t]{l}Payload:\\$\approx 2.5\text{kg}$\end{tabular}}}}%
  \end{picture}%
\endgroup%

    \caption{Unitree Go2 quadruped with approximately \SI{2.5}{\kg} of payload simulating measurement equipment during field tests on Vulcano, Italy. The model is updated at runtime.}
    \label{fig:title}
    \vspace{-0.4cm}
\end{figure}

Model-based control has driven significant advances in legged robotics in recent years. However, deploying these robots efficiently in many real-world applications—such as sample collection or measurement gathering (see~\autoref{fig:title}) in hostile environments or load carrying in industrial settings—requires the ability to operate with variable, unknown payloads. Traditional model-based control architectures rely on a fixed plant model and can only handle known payloads by manually incorporating their dynamic parameters into the equations of motion (EOM). To overcome this limitation, several approaches have been proposed, which we classify into five categories:
\begin{enumerate}
    \item model updates based on least-squares fitting~\cite{tournois_online_2017,ding_locomotion_2020}
    \item model updates based on tracking error evaluation~\cite{tournois_online_2017,ding_locomotion_2020,minniti_adaptive_2022,jin_unknown_2022}
    \item external force/disturbance estimation~\cite{minniti_adaptive_2022, elobaid_adaptive_2025, sombolestan_adaptive-force-based_2024, kang_external_2024, hawley_kalman_2018}
    \item machine learning (ML) based approaches~\cite{sun_online_2021,zeng_adaptive_2024, yao_adaptive_2023}
    \item model updates based on KF~\cite{evain_improving_2024, majeed_aerodynamic_2013, hong_vehicle_2013, hong_novel_2015, song_load_2020, joukov_constrained_2015}
\end{enumerate}

Fitting plant models to measurements taken from the real system using least-squares regression is a standard technique in system identification. Since model-based controllers such as MPC directly utilize the EOM, they are amenable to incorporating this approach online~\cite{marafioti_persistently_2014}, which is the core idea of the first category. In quadrupeds, this has been implemented for the online mass~\cite{ding_locomotion_2020} and COM~\cite{tournois_online_2017} estimation using a fitting method similar to RLS based on ground reaction forces (GRF). The update can only stabilize the controlled system if the estimates converge to the true values, for which sufficient excitation is crucial~\cite{bitmead_persistence_1984}. Furthermore, the estimation quality is highly susceptible to noise and biases in the system state and GRFs, which necessitates incorporating several heuristics~\cite{tournois_online_2017}. 

These limitations are addressed in the second category by estimating model parameters~\cite{tournois_online_2017,ding_locomotion_2020,minniti_adaptive_2022, jin_unknown_2022} such that the tracking error is minimized. In~\cite{minniti_adaptive_2022,jin_unknown_2022} the update law is validated using a Lyapunov stability analysis guaranteeing to stabilize the controlled system if it follows the dynamics of the Lyapunov function candidate \cite{slotine_adaptive_1987}. For this, computationally expensive constraints need to be added to the model-based controller. No information about the contact forces is needed but the performance is tied to the overall tracking capabilities of the system~\cite{tournois_online_2017}.

Another approach for legged robots to adapt to payloads is estimating externally applied forces or disturbances instead of changing model parameters. The results are fed to the controller and enable appropriate planning. They can again be estimated based on fitting to measurements~\cite{kang_external_2024,hawley_kalman_2018} or evaluating the tracking error~\cite{minniti_adaptive_2022,elobaid_adaptive_2025, sombolestan_adaptive-force-based_2024}. As the model uncertainties are treated as arbitrary disturbances this approach is more versatile than the previous approaches. The disadvantages described earlier also apply here.

The ML based methods aim to remove the effects of modeling nuances, by approximating error or disturbance models using ML techniques. The learned models are used either as residuals to remove the gap between the nominal model and the real world~\cite{sun_online_2021} or to estimate disturbances~\cite{zeng_adaptive_2024, yao_adaptive_2023}.

KFs are commonly used for state estimation but can also be utilized to estimate model parameters. In application areas such as satellite control~\cite{evain_improving_2024}, flight control~\cite{majeed_aerodynamic_2013}, automotive assistance systems~\cite{hong_vehicle_2013,wielitzka_joint_2015,hong_novel_2015}, and robot manipulators~\cite{song_load_2020,joukov_constrained_2015}, they have proven to be more reliable and resistant to noise than RLS in online and offline system identification. However, this method has not yet been utilized for the system identification or adaptive control of legged robots.

This paper presents a novel method for adaptive model-based control of a quadruped robot using Kalman filters. The benefit of our approach is that it is computationally cheap, and in contrast to the most other methods in the literature stated earlier, it also gives a direct measure of how reliable the current estimate is by outputting its covariance matrix. This enables reasoning, when to update the parameters. We find that the KF improves base height and orientation tracking and enables dynamic payload switching. It is more resistant to measurement noise and biases than a classical RLS method and by utilizing the estimation covariance we can reduce the number of heuristics needed for stable adaptation to a single threshold. We provide an open source implementation of our approach for the Unitree Go2 robot~\footnote{\href{https://github.com/dfki-ric-underactuated-lab/dfki-quad}{https://github.com/dfki-ric-underactuated-lab/dfki-quad}}. 

\paragraph*{Organization} The paper is structured as follows:~\autoref{sec:model} introduces the mathematical model of the robot followed by the description of the proposed KF and its implementation within a model-based control architecture~\cite{stark_benchmarking_2025} in~\autoref{sec:kf}. \autoref{sec:results} shows and discusses the experimental validation of the proposed method. Conclusions and future research directions are outlined in~\autoref{sec:dis}.

\section{Model for Parameter Estimation}
\label{sec:model}

\begin{figure}[t]
    \centering
    \vspace{0.3cm}
    \def\svgwidth{0.8\columnwidth}
    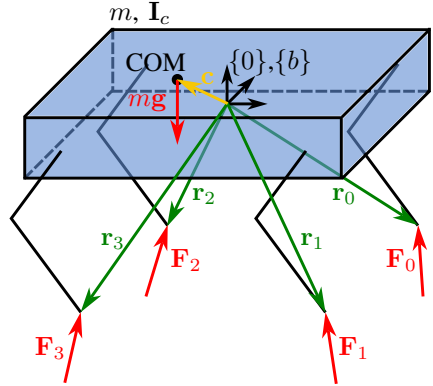
    \caption{Sketch of the quadruped model used for adaptive control. All mass is concentrated in the floating base. The model is defined in the inertial frame $\{0\}$ located at the geometric center of the floating base. The body frame $\{b\}$ is defined in the same location. The COM has an offset $\mathbf{c}$ from $\{0\}$. The inertia tensor $\mathbf{I}_c$ is defined about the COM in $\{b\}$. $\mathbf{F}_i$ are the GRFs and $\mathbf{r}_i$ the foot locations.}
    \label{fig:quad_sketch}
    \vspace{-0.4cm}
\end{figure}

In this section, the mathematical model used in the adaptive control of the quadruped is given.~\autoref{fig:quad_sketch} shows a sketch illustrating coordinate frames, acting forces, and geometric and mass-inertial quantities. The robot consists of a floating base in $\mathbb{SE}(3)$ and four legs with three degrees of freedom (DoF) each. In practice, the payload will most commonly be attached to the torso of the robot in an approximately rigid manner. Therefore, the floating base dynamics are of special interest, since the payload is encoded within its mass-inertial parameters. The legs are assumed to be massless and all mass is concentrated in the robot's floating base. This reduces the dimensionality of the estimation problem significantly as it reduces the equations of motion to a single rigid body actuated by four external forces:
\begin{align}
		&m(\dot{\mathbf{v}} + \mathbf{g}) = \sum_{i=0}^{3}{\mathbf{F}_i} \label{eq:EOM1}\\
		&\mathbf{R}^{T}(\mathbf{I}_c - m[\mathbf{c}]^2_\times)\mathbf{R}\dot{\boldsymbol{\omega}} + \mathbf{c}\times m\mathbf{g}= \sum_{i=0}^{3}(\mathbf{r}_i\times\mathbf{F}_i),
		\label{eq:EOM2}
\end{align}
$m$ is the robot's total mass, $\dot{\mathbf{v}}\in \mathbb{R}^3$ the linear acceleration, $\mathbf{g} = \begin{bmatrix}0, 0, g\end{bmatrix}^T\in \mathbb{R}^3$ the gravity vector and $\mathbf{F}_i\in \mathbb{R}^3$ the GRF at foot $i$. $\mathbf{R}\in \mathbb{R}^{3\times3}$ is the rotation from body frame $\{b\}$ to the inertial frame $\{0\}$ located at the body frame. $\mathbf{I}_c \in \mathbb{R}^{3\times3}$ is the rotational inertia about the COM, $\dot{\boldsymbol{\omega}} \in \mathbb{R}^3$ the angular acceleration in body frame, $\mathbf{c} \in \mathbb{R}^3$ the COM position and $\mathbf{r}_i$ the position of foot $i$. All vectors are represented in the inertial frame and $[\mathbf{c}]_\times$ denotes the skew-symmetric matrix constructed by $\mathbf{c}$. The term including $\dot{\boldsymbol{\omega}}$ can be neglected for the system identification process since the robot generally is not expected to perform high rotational acceleration movements. That further simplifies~\eqref{eq:EOM2} in the EOM to
\begin{equation}
    \mathbf{c}\times m\mathbf{g}= \sum_{i=0}^{3}(\mathbf{r}_i\times\mathbf{F}_i). \label{eq:EOM3}
\end{equation}
The foot positions $\mathbf{r}_i$ and GRFs $\mathbf{F}_i$ are obtained from leg kinematics and inverse dynamics:
\begin{align}
     & \mathbf{r}_i = \mathbf{f}_i(\mathbf{q}_i)\\ 
    & \mathbf{F}_i = (\mathbf{J}^T_i)^{-1}\boldsymbol{\tau}_i.\label{eq:inv_dyn}
\end{align}
$\mathbf{f}_i:\mathbb{R}^3\rightarrow\mathbb{R}^3$ denotes the kinematics of leg $i$ dependent on its joint positions $\mathbf{q}_i \in \mathbb{R}^3$. $\mathbf{J}_i \in \mathbb{R}^{3\times3}$ is the Jacobian matrix of leg $i$ and $\boldsymbol{\tau}_i \in \mathbb{R}^3$ its corresponding vector of joint torques.
Equations~\eqref{eq:EOM1} and~\eqref{eq:EOM3} contain the dynamic parameters to be estimated. Not all parameters are equally observable in every configuration. In quadrupedal walking with low roll and pitch angles, gravity excites the robot's mass and COM x- and y-components but not the z-component. Therefore it is assumed to be zero. The remaining parameters can be concatenated in the parameter vector:
\begin{align}
    &\boldsymbol{\pi} = \begin{bmatrix}
        m & \mathbf{h}^T
    \end{bmatrix}^T \in \mathbb{R}^{3}, \label{eq:param_vec}\\
    \text{with}\hspace{2mm} & \mathbf{h} = m\begin{bmatrix}c_x & c_y\end{bmatrix}^T \label{eq:com}
\end{align}
Since the EOM are affine in the parameters, they can be rearranged into
\begin{equation}
    \boldsymbol{\Phi} \boldsymbol{\pi}=\begin{bmatrix}
        \sum_{i=0}^{3}{\mathbf{F}_i} \\
        \sum_{i=0}^{3}(\mathbf{r}_i\times\mathbf{F}_i)
    \end{bmatrix}.\label{eq:regressor}
\end{equation}
$\boldsymbol{\Phi} \in \mathbb{R}^{6\times3}$ is the regressor matrix and is effectively the Jacobian of the EOM with respect to the parameter vector $\boldsymbol{\pi}$. It depends only on measurable quantities, i. e. robot pose and acceleration and the gravity vector. Each row of the regressor matrix corresponds to one DoF of the system, while each column shows the influence of one parameter on the system motion in that particular DoF.

\section{Kalman Filter for Adaptive Control} \label{sec:kf}

This section introduces the Kalman Filter (KF) for parameter identification and its application in adaptive control.
\subsection{Kalman Filter Design} \label{subsec:kf}
Consider the prediction step of a KF:
\begin{align}
    &\hat{\mathbf{x}}^-_{k} = \mathbf{A}\hat{\mathbf{x}}^+_{k-1} + \mathbf{B}\mathbf{u}_k \label{eq:pred_x}\\
    &\mathbf{P}^-_{k} = \mathbf{A}^T\mathbf{P}^+_{k-1}\mathbf{A} + \mathbf{Q}, \label{eq:pred_p}
\end{align}
where $\hat{\mathbf{x}}$ is the estimated state vector, $\mathbf{A}$ the Jacobian of the process model, $\mathbf{B}$ the mapping from input to state space and $\mathbf{u}$ the input vector. $\mathbf{P}$ is the estimation covariance matrix and $\mathbf{Q}$ the covariance matrix of the process noise. Superscript $-$ and $+$ denote quantities before and after the update step and subscript $k$ indicates a time step. When applying ~\eqref{eq:pred_x} and~\eqref{eq:pred_p} to the parameter identification problem, the estimated state vector $\hat{\mathbf{x}}$ becomes the estimated parameter vector $\hat{\boldsymbol{\pi}}$. Since there is no deterministic model of payload added to the system, the parameter vector is assumed to be stationary and without any inputs~\cite{joukov_constrained_2015}. A random walk is introduced via the process noise modeling unexpected parameter changes, i.e. added payload, via a Gaussian random distribution. This yields $\mathbf{A} = \mathbf{I}, \mathbf{B} = \mathbf{0}$ for the prediction step resulting in:
\begin{align}
    &\hat{\boldsymbol{\pi}}^-_k = \hat{\boldsymbol{\pi}}^+_{k-1}\label{eq:KF1}\\
    &\mathbf{P}^-_k = \mathbf{P}^+_{k-1} + \mathbf{Q},\label{eq:KF2}
\end{align}
with $\mathbf{P},\mathbf{Q}\in \mathbb{R}^{3\times3}$. 
The Kalman update step is
\begin{align}
    &\mathbf{K}_k = \mathbf{P}^-_k\mathbf{H}^T(\mathbf{H}\mathbf{P}^-_k\mathbf{H}^T + \mathbf{R})^{-1}\label{eq:upd_k}\\
    &\hat{\mathbf{x}}_k^+ = \hat{\mathbf{x}}_k^- + \mathbf{K}_k(\mathbf{z}_k-\mathbf{H}\mathbf{\hat{\mathbf{x}}_k^-})\label{eq:upd_x}\\
    &\mathbf{P}_k^+ = (\mathbf{I} - \mathbf{K_k\mathbf{H}})\mathbf{P}_k^-,\label{eq:upd_p}
\end{align}
where $\mathbf{K}$ is the Kalman gain, $\mathbf{H}$ the measurement model Jacobian, $\mathbf{R}$ the measurement covariance matrix, $\mathbf{z}$ the measurement vector and $\mathbf{I}$ the identity matrix. For the parameter estimation problem, the measurement model is given by the EOM. Therefore, $\mathbf{H}$ is given by the regressor matrix $\boldsymbol{\Phi}$. However, the regressor matrix is not constant over time as it depends on the current state, which is assumed to be perfectly known. The measurement vector $\mathbf{z}$ is given by the right-hand side of~\eqref{eq:regressor} which than yields for the update step:
\begin{align}
    &\mathbf{K}_k = \mathbf{P}^-_k\boldsymbol{\Phi}_k^T(\boldsymbol{\Phi}_k\mathbf{P}^-_k\boldsymbol{\Phi}_k^T + \mathbf{R})^{-1}\label{eq:KF3}\\
    &\hat{\boldsymbol{\pi}}_k^+ = \hat{\boldsymbol{\pi}}_k^- + \mathbf{K}_k(\mathbf{z}_k-\boldsymbol{\Phi}_k\mathbf{\hat{\boldsymbol{\pi}}_k^-})\label{eq:KF4}\\
    &\mathbf{P}_k^+ = (\mathbf{I} - \mathbf{K_k\boldsymbol{\Phi}_k})\mathbf{P}_k^-,\label{eq:KF5}
\end{align}
with $\mathbf{K}\in\mathbb{R}^{3\times6}$, $\mathbf{R}\in \mathbb{R}^{6\times6}$, $\mathbf{I}\in\mathbb{R}^{3\times3}$ and $\mathbf{z}\in\mathbb{R}^6$
\subsection{Adaptive Model-Based Control}\label{subsec:control}
\begin{figure}[tpb]
    \centering
    \vspace{0.3cm}
    \def\svgwidth{\columnwidth}
    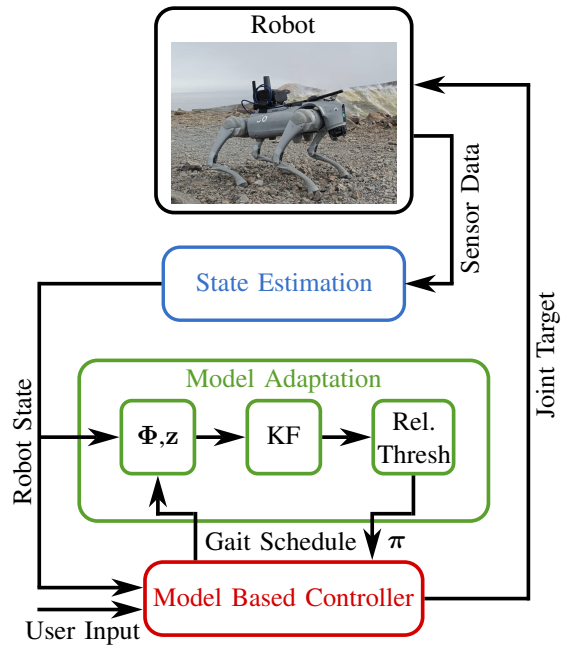
    \caption{Block diagram of the control architecture. A model-based controller sends joint targets to the robot according to the user input and robot state that is computed by the state estimation based on sensor feedback. The model adaptation sends its parameter estimate to all components at runtime.}
    \label{fig:control}
    \vspace{-0.4cm}
\end{figure}

The dynamic walking controller, shown in~\autoref{fig:control}, consists of four main components: the robot, state estimation, a model-based controller, and the model adaptation. 
The state estimation computes the robot's state from its proprioceptive sensors. 
The state consists of the floating base state including its accelerations, and foot contact forces. 
Based on the current estimated state and the user's input (target), the model-based controller plans a gait schedule and derives the joint commands for the robot's low-level joint control.
The model parameters are thereby updated online by the model adaptation.
In this work, the model-based controller~\cite{stark_benchmarking_2025} combines a heuristic-based gait sequencer, MPC, and whole-body control~\cite{mronga_arc-opt_2024}.

The model adaptation initially determines the regressor matrix $\boldsymbol{\Phi}$ and measurement vector $\mathbf{z}$ based on the estimated robot state, where the regressor matrix is obtained using the Pinnochio library \cite{carpentier_pinocchio_2019}. 
Since the state estimation determines the GRFs from~\eqref{eq:inv_dyn} they can be nonzero even without foot contact. 
To exclude faulty force measurements, both contact detection and gait scheduling are taken into account when computing $\mathbf{z}$. 
GRFs are only used if both a contact is measured and also scheduled. Otherwise, the respective force is set to zero. Once computed, $\boldsymbol{\Phi}_k$ and $\mathbf{z}_k$ are handed to the KF. A threshold is applied on the diagonal elements of the estimation covariance $\mathbf{P}$ to determine when parameter updates should be made based on their reliability. Finally, to isolate the floating base parameters, the contribution of the legs known from the robot’s digital model is subtracted.

\section{Results}\label{sec:results}
\begin{figure}[tpb]
    \centering
    \vspace{0.3cm}
    \includegraphics[]{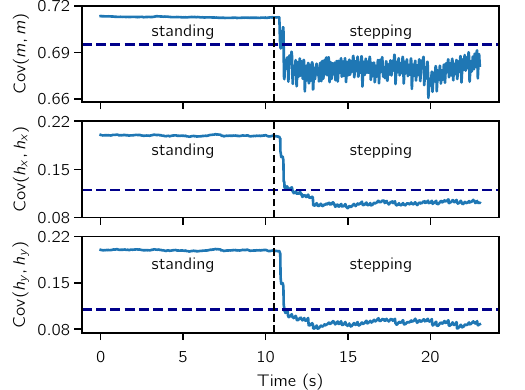}
    \caption{Diagonal elements of the estimation covariance matrix for KF. The robot stands still at first and begins stepping motions in trotting gait in place after approximately \SI{10}{\s}. The dark blue dashed lines represent the set thresholds on the covariance.}
    \label{fig:cov_stepping}
    \vspace{-0.4cm}
\end{figure}

The proposed approach is validated in hardware tests on a Unitree Go2 quadruped. Its total mass is \SI{16.21}{\kg} and the total COM offset is $c_x =$ \SI{8.8}{\mm}, and $c_y =$ \SI{0.0}{\mm}. MPC and model adaptation run at \SI{100}{\Hz} while the whole-body control runs at \SI{500}{\Hz}, and the state estimation at \SI{1000}{\Hz}. The behaviors of the following three dynamic walking controllers are compared:
\begin{enumerate}
    \item nominal control
    \item adaptive control with RLS parameter estimator
    \item adaptive control with KF parameter estimator.
\end{enumerate}
The RLS estimator is based on \cite{islam_recursive_2019} and the forgetting factor is set to $0.8$. The noise covariances of the KF are chosen as $\mathbf{Q} = \text{diag}(5 \cdot 10^{-3}, 5 \cdot 10^{-4}, 5 \cdot 10^{-4})$ and $\mathbf{R} = \text{diag}(10^3, 10^3, 10^4, 10^4, 10^4, 10^3)$. We choose $\begin{bmatrix} 0.695 & 0.12 & 0.11\end{bmatrix}$ as thresholds on the estimation covariances corresponding to $m$, $h_x$ and $h_y$ for the KF.

The thresholds are set because the GRFs from the state estimation are biased, especially during standing. Since the joint torques are measured via the motor currents, they do not show friction effects. This can lead to unwanted behavior when adapting the model online. During stepping motions, the joint friction is mostly overcome so that the calculated forces better resemble the applied forces, leading to a more reliable estimate at the expense of significantly higher noise.~\autoref{fig:cov_stepping} shows the diagonal elements of the estimation covariance matrix $\mathbf{P}$ from the KF while standing and stepping including the set thresholds. The covariances abruptly decrease when the robot starts stepping motions. This enables the model adaptation to send updates only when the force measurements are more reliable. For the RLS estimator, the covariances are not descriptive in that regard. We therefore omit the thresholds.

\subsection{Base Tracking under Load}
\begin{table}[t]
    \centering
    \vspace{0.3cm}
    \renewcommand{\arraystretch}{1.2}
    \caption{Mean absolute height ($\tilde{p}_z$), roll ($\tilde{\phi}$), and pitch ($\tilde{\theta}$) errors and their standard deviations of the evaluated controllers over in-place-stepping and walking experiments.}
    \begin{tabular}{@{}lllll@{}}
        \toprule
        payload (\SI{}{\g}) & controller & $\tilde{p}_z$ (\SI{}{\cm}) & $\tilde{\phi}$ (\SI{}{\deg}) & $\tilde{\theta}$ (\SI{}{\deg})\\
        \midrule
        \multirow{3}{*}{$0.0$}& nominal & $1.30 \pm 0.73$ & $0.47\pm 0.41$ & $1.15\pm0.96$ \\
        &RLS & $2.91\pm1.79$ & $1.20\pm1.38$ & $0.99\pm0.87$ \\
        &KF & $1.91\pm0.92$ & $0.67\pm0.60$ & $1.19\pm0.71$ \\
        \midrule
        \multirow{3}{*}{$1042.0$}&nominal & $2.59\pm0.87$ & $0.42\pm0.44$ & $1.62\pm1.06$ \\
        &RLS & $2.62\pm1.20$ & $1.09\pm1.59$ & $0.97\pm1.59$ \\
        &KF & $2.10\pm0.78$ & $0.95\pm0.77$ & $1.09\pm0.84$ \\
        \midrule
        \multirow{3}{*}{$2544.5$}&nominal & $3.43\pm1.31$ &$0.42\pm0.44$ & $1.63\pm1.25$\\
        &RLS & $3.73\pm2.09$ & $1.35\pm1.53$ & $1.22\pm0.98$\\
        &KF & $1.84\pm0.68$ & $1.18\pm0.89.$ & $1.01\pm0.77$\\
        \midrule
        \multirow{3}{*}{$5105.5$}&nominal & $4.92\pm1.81$ & $2.23\pm1.61$ & $1.60\pm1.42$ \\
        &RLS &  $4.55\pm2.35$ & $2.48\pm2.47$ & $1.41\pm0.84$ \\
        &KF & $1.89\pm0.72$ & $1.40\pm1.12$ & $1.05\pm0.80$ \\
        \bottomrule
    \end{tabular}
    \label{tab:base_tracking}
    \vspace{-0.5cm}
\end{table}
We evaluate the base tracking capabilities of the controllers for the following scenarios:
\begin{enumerate}
    \item The robot stands up from a lying position and stands still for approximately \SI{5}{\s}. Then it steps in place in trotting gait for another \SI{20}{\s}.
    \item The robot walks approximately \SI{7}{\m} in trotting gait, turns around and walks back to the starting position.
\end{enumerate}
Each scenario is repeated with either \SI{0.0}{\g}, \SI{1042.0}{\g}, \SI{2544.5}{\g} or \SI{5105.5}{\g} of payload attached to the robot's torso close to its center via tensioning straps. The target base height is set to \SI{0.3}{m}, and target roll and pitch angles to \SI{0.0}{\deg} over all experiments. No user inputs are applied during the first scenario. In the second scenario, an operator steers the robot between marked positions. To keep the comparison fair, no modifications to the hyper-parameters of the model-based controller (MPC, WBC, gait sequencer) were applied in all these scenarios.
\begin{figure*}[t]
    \centering
    \vspace{0.3cm}
    \includegraphics[]{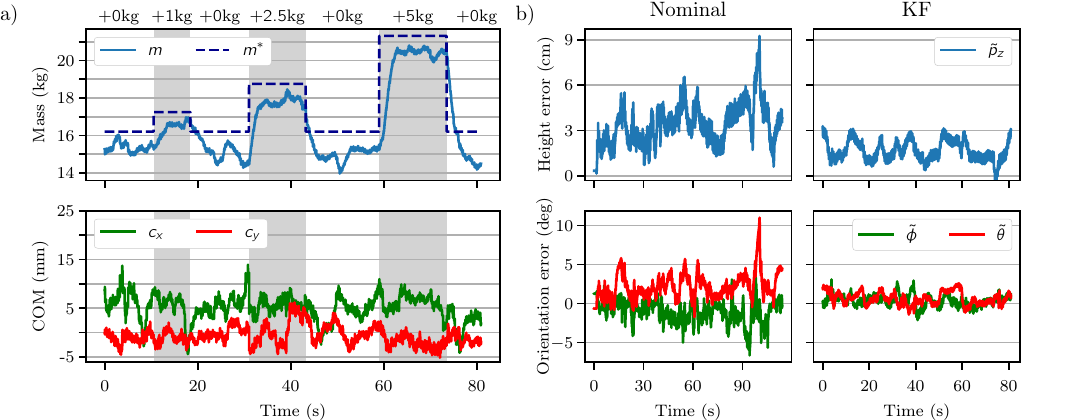}
    \caption{a) Estimated total mass with ground truth (top) and estimated COM (bottom) during dynamic payload switching experiment. The original robot weight is \SI{16.21}{\kg} and the COM offsets are $c_x =$ \SI{8.8}{\mm}, and $c_y =$ \SI{0.0}{\mm}. The time intervals where a specific amount of payload (displayed above the upper graph) is attached to the robot are highlighted in grey. b) Height and orientation error of nominal controller (left) and adaptive controller with KF (right) over dynamic weight switching experiment.}
    \label{fig:weight_switching}
    \vspace{-0.5cm}
\end{figure*}

~\autoref{tab:base_tracking} shows the mean absolute base position error in the z-axis $\tilde{p}_z$, mean absolute roll error $\tilde{\phi}$, mean absolute pitch error $\tilde{\theta}$, and their respective sample standard deviations over the two scenarios combined. The base height measurements are computed from the leg kinematics while the orientation measurements are computed by the state estimation from IMU data and leg odometry. The KF outperforms RLS and the nominal case for most payloads in the base height tracking. With payloads attached, it reduces the mean error and standard deviation. The only exception is without any payload. There the KF increases the mean height error and standard deviation by \SI{0.6}{\cm} and \SI{0.19}{\cm} compared to the nominal case. For all payloads, the KF keeps the height error approximately constant. RLS does not improve the base height tracking. The roll and pitch tracking performances of the nominal controller are only marginally impacted by the payload. The KF slightly reduces the roll error for a payload of \SI{5105.5}{\g}, and the pitch error for all payloads. 
The RLS results on orientation error are inconclusive.

\subsection{Dynamic Payload Switching}
The behavior of the KF is further evaluated regarding its adaptability towards changing payloads. For that the robot steps in place while the payload is changed successively.

~\autoref{fig:weight_switching} a) shows the estimated mass $m$ the corresponding ground truth $m^\ast$ and estimated COM x and y-components $c_x$ and $c_y$ over the experiment time. The COM estimates are presented without ground truth data as it cannot be accurately obtained in our experimental setups. The mass estimate fluctuates by approximately $\pm$\SI{500}{\g} around a mean value. Larger deviations occur when the robot performs abrupt movements to compensate for drift. There is also a bias of approximately \SI{1}{\kg} present. When payload is added the estimate quickly rises by the correct amount. Upon weight removal, the estimate converges to a value lower than the original mass. The estimate of $c_x$ fluctuates by approximately $\pm$\SI{5}{\mm} while $c_y$ does so at a lower degree around $\pm$\SI{2.5}{\mm}. The weight changes are not observable in the COM estimation.~\autoref{fig:weight_switching} b) shows the tracking errors of the nominal and adaptive controller over the same experiment. The adaptation improves the tracking of height and orientation maintaining a more consistent error across all payloads.

\subsection{Discussion}
The experiments demonstrate that online system identification using a KF enhances the height tracking performance of the model-based controller. During dynamic payload changes it also reduces the orientation error and helps maintain a more constant level. However, in the absence of a payload, performance deteriorates, likely due to the bias in the mass estimation. This bias is, in turn, attributed to biased force measurements. When payload is present, any mass estimation closer to the true value than in the nominal case can aid in compensation, even if it remains below the true mass. Conversely, without a payload, the nominal model more accurately represents the system than a biased estimated model, leading to degraded tracking performance. This issue could be mitigated by disregarding estimates within a certain range around the nominal parameters or by improving force measurement accuracy.

The bad performance of the RLS estimator is most likely because its error covariances do not enable meaningful thresholding to distinguish if the estimate is reliable. Without other logic, it is then subject to biases in the contact forces, that are caused by e.g. joint friction while standing still. During stepping motions, it is also not as resistant to the large noise as the KF.

Although~\autoref{fig:weight_switching} b) shows an improvement in the orientation tracking it is not that obvious in~\autoref{tab:base_tracking}. Also, the estimates in~\autoref{fig:weight_switching} a) do not show large changes in the COM. This is probably caused by the placement of the payload during the experiments. When strapped tightly to the torso the weights were positioned close to the robot center. Therefore large deviations from the nominal COM are not expected. For better evaluation, a different experiment setup is needed that allows a larger range of positioning options for the payload. This way big COM changes for the controller to react to could be provoked.

The parameter estimates in~\autoref{fig:weight_switching} a) show strong noise. This is likely caused by the impacts of the footsteps that introduce noise to the contact forces and the regressor matrix, which is assumed to be perfectly known. Especially the induced vibrations influence the acceleration signals and lead to problems in the estimation. This can be addressed by improving state estimation, formulating the problem in a way that does not use accelerations, or making the method more robust to uncertainties in the regressor matrix.
\vspace{-0.1cm}
\section{CONCLUSION AND FUTURE WORK}\label{sec:dis}
We presented a KF based formulation that identifies the mass and COM of a quadruped at runtime. Under payload, it reduces the tracking error and can keep it constant even for payloads as large as $30\%$ of the robot's body weight. While it is still susceptible to biases and noise, its influence is reduced compared to RLS estimation. The estimation covariance enables reasoning about the reliability of the algorithm output. This property is novel among state-of-the-art methods for adaptive model-based control of quadrupeds. By utilizing that, we were able to reduce the number of heuristics needed for stable online estimation to a single threshold. Further, the information could be used in the controller or state estimation, to make them more robust to model uncertainties. Unlike tracking error-based methods that often introduce non-linear constraints to the optimization problem, the algorithm avoids computationally expensive operations. However, this benefit comes at the cost of stability guarantees.
Future research includes extending the method to additionally estimate the second moments of inertia with physical consistency constraints~\cite{wensing_linear_2018} by introducing a constrained KF~\cite{simon_kalman_2024}. Since the inertia tensor is not always observable, additional considerations about observability and persistency of excitation would need to be included. Furthermore, the method can be improved regarding its resistance to noise, especially in the regressor matrix. Joint friction could also be estimated online to reduce the GRF bias.

\vspace{-0.1cm}
\bibliographystyle{IEEEtran}
\bibliography{KFforAdaptiveMPC}

\end{document}